\documentclass[AMS,Times2COL]{WileyNJDv5}
\usepackage{graphicx} 
\usepackage{pifont}
\usepackage{bm}
\articletype{Research Article}%
\received{Date Month Year}
\revised{Date Month Year}
\accepted{Date Month Year}
\journal{Journal}
\volume{00}
\copyyear{2023}
\startpage{1}
\usepackage{hyperref}
\usepackage{orcidlink}

\hypersetup{
            colorlinks=true,
            linkcolor=blue,
            anchorcolor=blue,
            citecolor=blue
            }
\usepackage{balance}

\begin{document}

\title{MSLAU-Net: A Hybrid CNN-Transformer Network for Medical Image Segmentation}

\author[1]{Libin Lan}
\author[1]{Yanxin Li}
\author[2]{Xiaojuan Liu}
\author[3]{Juan Zhou}
\author[2]{Jianxun Zhang}
\author[4,5,6]{Nannan Huang}
\author[7]{Yudong Zhang}

\authormark{LAN \textsc{et al.}}
\titlemark{MSLAU-NET: A HYBRID CNN-TRANSFORMER NETWORK FOR MEDICAL IMAGE SEGMENTATION}

\address[1]{\orgdiv{College of Computer Science and Engineering}, \orgname{Chongqing University of Technology}, \orgaddress{\state{Chongqing}, \country{China}}}
\address[2]{\orgdiv{College of Artificial Intelligence}, \orgname{Chongqing University of Technology}, \orgaddress{\state{Chongqing}, \country{China}}}
\address[3]{\orgdiv{Department of Pharmacy}, \orgname{the Second Affiliated Hospital of Army Military Medical University}, \orgaddress{\state{Chongqing}, \country{China}}}
\address[4]{\orgdiv{Department of Prosthodontics}, \orgname{the Affiliated Stomatological Hospital of Chongqing Medical University}, \orgaddress{\state{Chongqing}, \country{China}}}
\address[5]{\orgname{Chongqing Key Laboratory of Oral Diseases}, \orgaddress{\state{Chongqing}, \country{China}}}
\address[6]{\orgname{Chongqing Municipal Key Laboratory of Oral Biomedical Engineering of Higher Education}, \orgaddress{\state{Chongqing}, \country{China}}}
\address[7]{\orgdiv{School of Computer Science and Engineering}, \orgname{Southeast University}, \orgaddress{\state{Nanjing}, \country{China}}}

\corres{Libin Lan, College of Computer Science and Engineering, Chongqing University of Technology, Chongqing 400054, China. \email{lanlbn@cqut.edu.cn}}

\abstract[Abstract]{Accurate medical image segmentation allows for the precise delineation of anatomical structures and pathological regions, which is essential for treatment planning, surgical navigation, and disease monitoring.
Both CNN-based and Transformer-based methods have achieved remarkable success in medical image segmentation tasks. However, CNN-based methods struggle to effectively capture global contextual information due to the inherent limitations of convolution operations. Meanwhile, Transformer-based methods suffer from insufficient local feature modeling and face challenges related to the high computational complexity caused by the self-attention mechanism. To address these limitations, we propose a novel hybrid CNN-Transformer architecture, named MSLAU-Net, which integrates the strengths of both paradigms.
The proposed MSLAU-Net incorporates two key ideas. First, it introduces Multi-Scale Linear Attention, designed to efficiently extract multi-scale features from medical images while modeling long-range dependencies with low computational complexity. Second, it adopts a top-down feature aggregation mechanism, which performs multi-level feature aggregation and restores spatial resolution using a lightweight structure. 
Extensive experiments conducted on benchmark datasets covering three imaging modalities demonstrate that the proposed MSLAU-Net outperforms other state-of-the-art methods on nearly all evaluation metrics, validating the superiority, effectiveness, and robustness of our approach.
Our code is available at \url{https://github.com/Monsoon49/MSLAU-Net}.
}


\maketitle

\section{Introduction}
\label{sec1}
Medical image analysis plays a pivotal role in computer-aided diagnosis and modern medical practice \cite{chen2021transunet,cao2022swin}. Central to this process is robust and efficient medical image segmentation, which provides the foundation for accurate diagnosis, treatment planning, and disease monitoring \cite{du2022swinpa,srivastava2021msrf}. Advanced computational methods, such as convolutional neural networks (CNNs) and transformers, enhance segmentation reliability by ensuring consistent extraction of clinically relevant insights from complex data. These techniques not only improve diagnostic accuracy but also offer critical applications in tracking disease progression and guiding preventive care strategies.

In recent years, CNNs have achieved significant success in medical image segmentation. The fully convolutional network (FCN) \cite{long2015fully} pioneers this advancement, but limitations in detail preservation and contextual understanding restrict its broader applicability. To address these challenges, U-Net \cite{ronneberger2015u} introduces an encoder-decoder architecture with skip connections, thereby effectively retaining finer details and improving accuracy. This design is particularly well-suited for medical image segmentation tasks, enabling the precise identification and localization of fine structures within target regions. Building on the achievements of U-Net, numerous U-shaped architectures have been developed, such as 3D U-Net \cite{cciccek20163d}, V-Net \cite{milletari2016v}, U-Net++ \cite{zhou2018unet++}, and DenseUNet \cite{li2018h}. Despite their proven success in medical image segmentation, CNN-based methods are inherently constrained by the fixed receptive fields of convolutional kernels, hindering their capacity to capture spatial long-range dependencies and global semantic information. To address this limitation, several studies have proposed employing dilated convolutions \cite{chen2018encoder,jiang2020hdcb,chen2017deeplab,gu2019net}, attention mechanisms \cite{schlemper2019attention,wang2018non}, and image pyramid frameworks \cite{zhao2017pyramid,peng2020semantic}. Despite these efforts, effectively capturing long-range spatial dependencies remains a persistent challenge for such methods.

Inspired by the success of transformers in natural language processing (NLP) \cite{vaswani2017attention} and computer vision \cite{dosovitskiy2020image}, the self-attention mechanism has garnered significant attention for its ability to capture global dependencies and emphasize critical features in medical imaging. Several representative studies, such as nnFormer \cite{zhou2023nnformer}, UTNet \cite{gao2021utnet}, TransUNet \cite{chen2021transunet}, HiFormer \cite{heidari2023hiformer} and MISSFormer \cite{huang2021missformer}, have focused on integrating transformers into medical image segmentation tasks.
However, vanilla self-attention mechanisms suffer from high computational complexity and memory demands, limiting their broad applicability and efficiency in medical image segmentation tasks. 

To address these challenges, various modifications have been proposed. Among them, sparse attention mechanisms are frequently proposed as an effective solution to mitigate these constraints, with notable methods including Swin-Unet \cite{cao2022swin}, Gated Axial UNet (MedT) \cite{valanarasu2021medical}, BRAU-Net \cite{cai2024pubic}, and BRAU-Net++ \cite{lan2024brau}.
Despite their effectiveness, sparse attention typically restricts the receptive field by focusing only on a small portion of the input sequence or elements within a predefined window. This limitation risks neglecting critical long-range positions, resulting in an incomplete representation of the overall context. 

In pursuit of both efficiency and a global receptive field, alternative linear-complexity models have emerged. State space models (e.g., Mamba \cite{gu2024mamba}) have shown promise but are inherently designed for 1D sequences, making their adaptation to 2D images non-trivial. In contrast, linear attention, rooted in the transformer framework, offers a more direct and architecturally seamless pathway for global contextual modeling in visual tasks. Specifically, linear attention approximates the original Softmax function using feature mapping or mathematical transformations and exploits the associativity of matrix multiplication to reorder computations from ($\bm{Q}\bm{K})\bm{V}$ to $\bm{Q}(\bm{K}\bm{V})$ \cite{bolya2022hydra,han2023flatten,han2024agent}. By doing so, the computational complexity is effectively reduced to $\mathcal{O}(N)$.
However, to the best of our knowledge, the application of linear attention to medical image segmentation remains underexplored, motivating us to investigate its potential in this domain.

The integration of attention mechanism with feature fusion has emerged as a pivotal strategy for developing lightweight yet high-performance models in medical image analysis. Exemplifying this trend, Li et al. \cite{wang2023lightweight} proposed an attention mechanism feature fusion model for gastric lesion classification, which achieves effective feature integration and enhancement within a CNN-based framework. Their work validates the effectiveness of a synergistic design that combines attention mechanism with feature fusion.

Motivated by the core ideas discussed above, we propose to incorporate feature fusion directly into the design of linear attention. In fact, multi-scale learning \cite{chen2017deeplab,zhao2017pyramid,rahman2024emcad} and global receptive field \cite{xie2021segformer} have been widely recognized as critical factors for enhancing model segmentation performance \cite{cai2023efficientvit}.
While linear attention inherently satisfies the need for global contextual understanding, most existing linear attention approaches operate at a single scale, failing to fully exploit multi-scale information. However, it is well-known that multi-scale information is particularly important for medical image segmentation tasks, as it encapsulates both fine-grained details and high-level semantic patterns. This capability is essential for handling significant variations between different tissues and lesion areas, ranging from microscopic cellular structures to macroscopic organ levels. To this end, we propose a novel \textbf{M}ulti-\textbf{S}cale \textbf{L}inear \textbf{A}ttention (MSLA) module, which integrates depth-wise convolutions at multiple scales to extract hierarchical features from input data and employs linear attention to aggregate cross-scale global context. This design simultaneously achieves scale-specific feature refinement and efficient global contextual modeling, ensuring comprehensive representation learning.

Using MSLA as a foundational building module, we design transformer-based Global Feature Extraction (GFE) blocks. Combining these GFE blocks with CNN-based Local Feature Extraction (LFE) blocks, adapted from the Local UniFormer block \cite{li2023uniformer}, we construct a four-stage encoder. In the first two stages, we utilize LFE blocks for local feature modeling, while in the last two stages, we employ GFE blocks (see Fig. \ref{fig3:gfe}) to capture global features. At its core, the GFE block integrates the MSLA module to enable multi-scale feature extraction and computationally efficient linear attention, ensuring robust hierarchical representation learning.
For the decoder, instead of adopting conventional symmetric U-shaped architectures, we incorporate a top-down multi-level feature aggregation mechanism. This mechanism fuses features from the encoder stages using lightweight convolutional layers and bilinear upsampling operations, progressively enriching spatial details and semantic coherence during decoding. Built upon this encoder-decoder design, we propose MSLAU-Net, a hybrid architecture that synergizes the strengths of CNNs and transformers for efficient medical image segmentation.

Our main contributions are three-fold:
\begin{itemize}
\item We propose a novel MSLA module designed in a parallel architecture to fully leverage the strengths of CNNs for capturing multi-scale low-level fine-grained details and linear attention for modeling long-range dependencies. This design enables the model to simultaneously benefit from the local feature extraction and the global context understanding while maintaining low computational complexity, which is essential for improving both the computational efficiency and segmentation performance of the model.
\item We design a top-down aggregation mechanism on the decoder side to aggregate multi-level features from the corresponding stages of the encoder. All aggregated features are then upsampled to recover the original spatial resolution. Building upon this aggregation mechanism and the proposed MSLA module, we introduce MSLAU-Net, a hybrid CNN-Transformer encoder-decoder architecture that adopts an asymmetric design instead of the conventional symmetric U-shaped networks, enabling the model to efficiently segment target regions.
\item We extensively evaluate MSLAU-Net on three benchmark medical image datasets: Synapse Multi-Organ Segmentation \cite{landman2015miccai}, Automated Cardiac Diagnosis Challenge \cite{bernard2018deep}, and CVC-ClinicDB \cite{bernal2015wm}. Experimental results demonstrate that MSLAU-Net achieves state-of-the-art performance with excellent generalization and robustness across diverse medical imaging tasks.
\end{itemize}

The remainder of this paper is organized as follows. Section \ref{sec:relatedwork} delves into the findings and limitations derived from previous relevant work. Section \ref{sec:method} illuminates our model’s design and architecture. Section \ref{sec:experiments} presents experimental results, visualizes MSLA attention maps, and provides an in-depth analysis of performance. Section \ref{sec:conclusion} summarizes our findings, points out the limitations of the work, and outlines directions for future research.

\section{Related work}
\label{sec:relatedwork}
Our work builds upon both CNN-based and transformer-based methods for medical image segmentation. A key contribution is the improvement of the transformer's core building module, i.e., the attention mechanism by proposing a novel linear attention module aimed at enhancing computational efficiency. We therefore review relevant work in CNN and transformer segmentation methods, linear attention, as well as modern sequence modeling techniques.

\subsection{Medical Image Segmentation}
\textbf{CNN-Based Methods}. Initially, U-shaped networks developed for medical image segmentation utilized CNN operations to achieve precise results. U-Net \cite{ronneberger2015u} was the first to introduce this distinctive U-shaped architecture, designed to effectively obtain multi-scale information and fuse features (via skip connections), which makes it particularly well-suited for preserving local fine-grained details and capturing global semantic information. 

Since then, several variants have been proposed to further improve performance \cite{zhou2018unet++, huang2020unet,cciccek20163d}. For instance, U-Net++ \cite{zhou2018unet++} introduces nested and dense skip connections to reduce the semantic gap between the encoder and decoder. UNet 3+ \cite{huang2020unet} optimizes full-scale skip connections and incorporates deep supervision to boost segmentation accuracy. Additionally, 3D-Unet \cite{cciccek20163d} extends the original U-Net design by incorporating 3D convolutions into its architecture, which can effectively process volumetric medical data, such as CT and MRI scans. 

Owing to the powerful local feature representation capabilities of CNNs, CNN-based U-shaped architectures have demonstrated exceptional effectiveness\iffalse outstanding performance remarkable results \fi in medical image segmentation. However, convolutional operations are inherently limited in their ability to capture long-range dependencies. To address this limitation, we propose incorporating transformers into our model architecture to compensate for the shortcomings of CNNs. Specifically, we replace CNNs with transformers in the deeper encoding stages of our model.

\textbf{Transformer-Based Methods}. Motivated by the success of transformers in NLP \cite{vaswani2017attention}, there has recently been growing interest in applying transformers to computer vision tasks \cite{dosovitskiy2020image,zhu2023biformer,li2023uniformer}, which has proven effective due to its ability to model long-range dependencies. As a result, more approaches have emerged that aim to leverage transformers for medical image analysis, particularly for medical image segmentation, which requires a comprehensive understanding of both structural details and global context \cite{zhou2023nnformer,gao2021utnet,heidari2023hiformer,huang2021missformer,cai2024pubic,hatamizadeh2022unetr}. 

Transformer-based representative works in medical image segmentation include TransUNet \cite{chen2021transunet}, which integrates the merits of CNNs and transformers for local feature extraction and global contextual modeling, respectively, within a U-shaped architecture; Swin-Unet \cite{cao2022swin}, which adopts a pure transformer architecture based on Swin Transformer \cite{liu2021swin}, enabling efficient computation; and BRAU-Net++ \cite{lan2024brau}, which innovatively introduces dynamic sparse attention into a hybrid CNN-Transformer architecture and redesigns skip connection by using channel-spatial attention mechanism. 

Despite these advancements, existing transformer-based methods often rely on Softmax attention or sparse attention mechanisms, which still present challenges such as high computational costs and limited receptive field sizes. In contrast to these approaches, we propose using linear attention to achieve lower computational complexity and provide a global receptive field, so as to overcome some of the limitations associated with Softmax and sparse attention mechanisms.

\subsection{Linear Attention and Modern Sequence Modeling}
\textbf{Linear Attention}. Linear attention reformulates the self-attention mechanism by employing kernel functions instead of Softmax function, reducing complexity to $\mathcal{O}(N)$ via the associative property of matrix multiplication, while maintaining a global receptive field \cite{katharopoulos2020transformers}. The pioneering work \cite{katharopoulos2020transformers} introduces a feature mapping $\phi$ on $\bm{Q}$ and $\bm{K}$, but simple feature mappings often lead to performance degradation. Subsequent methods, including Efficient Attention \cite{shen2021efficient}, Hydra Attention \cite{bolya2022hydra}, Focused Linear Attention \cite{han2023flatten}, and Agent Attention \cite{han2024agent}, have been proposed to enhance the approximation through various designs. 

While these methods have proven effective, their capacity to model complex patterns remains limited when compared to Softmax attention. Moreover, they operate exclusively on a single scale, neglecting the potential benefits of multi-scale feature exploration. The importance of such exploration is underscored by Alam et al. \cite{alam2017alzheimer}, who demonstrated the advantages of composite feature representations in neurological image analysis by integrating multi-scale features through multi-kernel learning. EfficientVit \cite{cai2023efficientvit} represents a relevant endeavor to incorporate multi-scale learning into linear attention via depth-wise convolutions. Nevertheless, its reliance on small-kernel convolutions constrains comprehensive feature extraction. Our work addresses this gap by introducing the \textbf{M}ulti-\textbf{S}cale \textbf{L}inear \textbf{A}ttention (MSLA) module, specifically designed to learn comprehensive multi-scale features and significantly enhance the expressive power of linear attention.

\textbf{Modern Sequence Modeling}. Beyond linear attention, state space models (SSMs) have recently emerged as competitive alternatives for long-range dependency modeling. The Mamba architecture \cite{gu2024mamba} introduces data-dependent selection mechanisms into structured state space models, achieving linear computational complexity while maintaining strong performance. This has inspired several medical image segmentation approaches, including VM-UNet \cite{ruan2024vm} which implements a Mamba-based U-shaped architecture utilizing Visual State Space (VSS) blocks for efficient long-range modeling.

Although Mamba shares the goal of efficient long-range modeling with linear attention, the two approaches are fundamentally distinct. Mamba relies on state space models and selective scanning mechanisms, whereas linear attention is derived from a kernel-based reformulation of the standard attention. This foundational difference poses a significant challenge for Mamba in medical image segmentation: its inherently 1D sequential nature requires complex adaptation strategies (e.g., cross-scanning) to handle 2D spatial relationships, which can disrupt the natural locality and coherence in images. In contrast, our MSLA module is natively designed for 2D medical images, inherently preserving spatial structure while leveraging the efficiency of linear attention, offering a more direct and effective solution for this domain.

\section{Method}
\label{sec:method}
In this section, we detail the proposed approach. First, we provide a concise summary of Efficient Attention. Next, we elaborate on the \textbf{M}ulti-\textbf{S}cale \textbf{L}inear \textbf{A}ttention (MSLA) module. Then, we introduce two key components: an encoder built on the MSLA module and a decoder designed to better preserve low-level spatial details while enhancing high-level semantic information. Finally, we specify the overall architecture of the proposed MSLAU-Net and its associated loss function.

\subsection{Preliminaries}
Given the input $ \bm{X} \in {\mathbb{R}^{N \times C}}$ with $N$ tokens, the general form of self-attention can be written as follows in each head \cite{han2023flatten}:
\begin{equation} \label{eq:general_attn}
    \begin{split}
        \bm{Q}=\bm{X}\bm{W}_q, \bm{K}=\bm{X}\bm{W}_k, \bm{V}=\bm{X}\bm{W}_v, \\
        \bm{O}_i=\sum_{j=1}^{N}\ \frac{{\rm Sim}{\left(\bm{Q}_i,\bm{K}_j\right)}}{\sum_{j=1}^{N}\ {\rm Sim}{\left(\bm{Q}_i,\bm{K}_j\right)}}\bm{V}_j,
    \end{split}
\end{equation}
where $ \bm{W}_q, \bm{W}_k, \bm{W}_v\!\in\!\mathbb{R}^{C \times d} $ are learnable linear projection matrices, $C$ and $d$ are the channel dimension of module and each head, and $ {\rm Sim}{\left(\cdot,\cdot \right)} $ denotes the similarity function.

In modern vision transformer architectures, the most widely adopted form of attention is Softmax attention \cite{vaswani2017attention}, which operates by applying a Softmax function to the scaled dot-product of queries and keys. Its similarity function is defined as 
$ \text{Sim}(\bm{Q}, \bm{K}) = \exp({\bm{Q}\bm{K}^{\top}}/{\sqrt{d}}). $
Softmax attention requires computing the similarity between all query-key pairs, leading to the computational complexity of $\mathcal{O}(N^2)$.

To address this issue, linear attention has been proposed as an alternative, significantly reducing the complexity. Specifically, linear attention reformulates the attention mechanism by leveraging mapping function $\phi(\cdot)$, resulting in the similarity function formulated as 
${\rm Sim}\left(\bm{Q},\bm{K}\right)=\phi(\bm{Q})\phi(\bm{K})^{\top}$.

Efficient Attention, a type of linear attention, maintains similar representational power to Softmax attention while achieving linear computational complexity \cite{shen2021efficient}. Mathematically, Efficient Attention can be expressed as follows:
\begin{equation}
    {\rm EfficientAtt}(\bm{Q},\bm{K},\bm{V})\!=\!{\rm Sim}(\bm{Q},\bm{K})\bm{V} \!=\!\phi_q(\bm{Q}){\phi_k(\bm{K})}^{\top}\bm{V}\!.
\end{equation}
The mapping functions for Efficient Attention are defined as:
\begin{equation}
    \begin{split}
        {\phi}_{q}(\bm{Q})={\rm \sigma}_{row}(\bm{Q}), \\
        {\phi}_{k}(\bm{K})={\rm \sigma}_{col}(\bm{K}),
    \end{split}
\end{equation}
where ${ \rm \sigma}_{row}(\bm{Q}), {\rm \sigma}_{col}(\bm{K})$ denote the application of the softmax function along each row of query matrix and each column of key matrix, respectively. Based on the associative property of matrix multiplication, the computation order can be changed from $(\phi_{q}(\bm{Q})\phi_{k}(\bm{K})^{\top})\bm{V}$ to $\phi_{q}(\bm{Q})(\phi_{k}(\bm{K})^{\top}\bm{V})$. By doing so, the computational complexity of Efficient Attention is reduced to $\mathcal{O}(N)$.

\subsection{Multi-Scale Linear Attention}
Our Multi-Scale Linear Attention (MSLA) consists of two main operational processes: Multi-Scale Feature Extraction and Linear Attention Computation. The former captures multi-scale local structural details to enhance segmentation performance. The latter leverages linear attention with an approximated global receptive field, similar to that of Softmax attention, to model long-range dependencies while improving computational efficiency. A detailed explanation is provided below.

\subsubsection{Multi-Scale Features Extraction}
As illustrated in Fig. \ref{fig1:msla}, we first reshape the input tokens into a feature map $\bm{X} \in {\mathbb{R}^{\sqrt{N} \times \sqrt{N} \times C}}$, and then split it into four parts along the channel dimension $C$:
\begin{figure*}[htbp]
\centering
\includegraphics[width=1.0\linewidth, keepaspectratio]{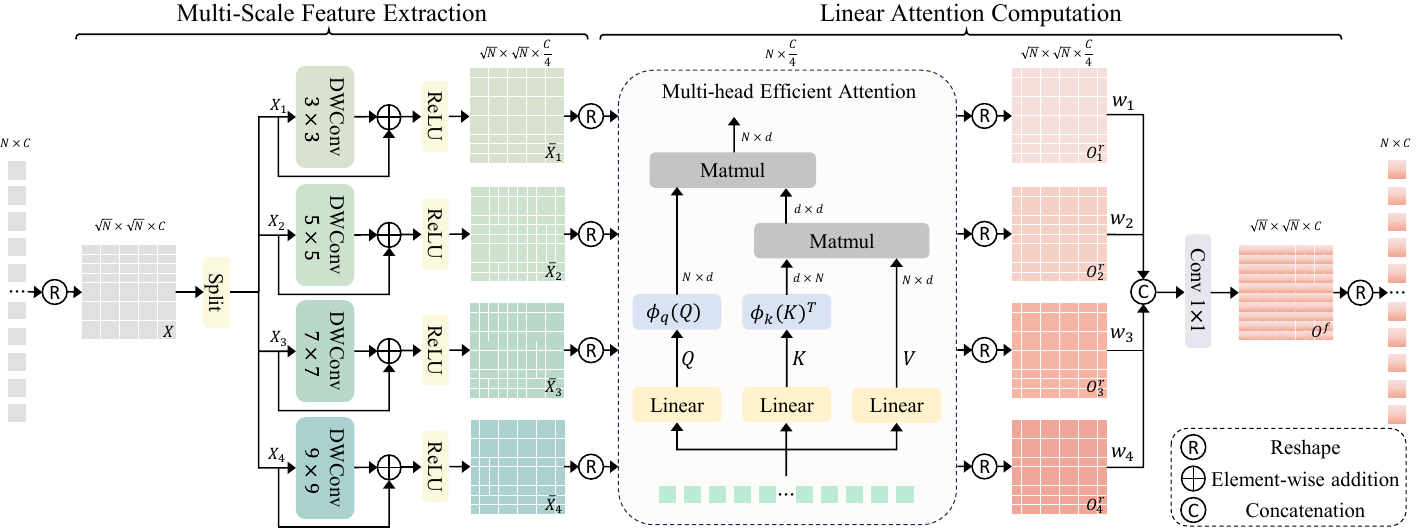}
\caption{Details of Multi-Scale Linear Attention. The MSLA module is designed in parallel to take full advantage of CNNs for capturing multi-scale features and linear attention for modeling long-range dependencies. The input feature map is first divided into four parts along the channel dimension. Each part is then processed through depth-wise convolution with different kernel sizes (3$\times$3, 5$\times$5, 7$\times$7, and 9$\times$9) to extract multi-scale features. Next, linear attention, i.e., Efficient Attention, is applied to the multi-scale features to model long-rage dependences. Finally, the resulting outputs are fused using a 1$\times$1 convolution.}
\label{fig1:msla}
\end{figure*}
\begin{equation}
    \bm{X}_1, \bm{X}_2, \bm{X}_3, \bm{X}_4 = \text{Split}\left(\bm{X}, \frac{C}{4}\right).
\end{equation}
After reshaping and splitting, these features are then fed into four parallel depth-wise convolution branches designed to explore multi-scale representations using 3$\times$3, 5$\times$5, 7$\times$7, and 9$\times$9 kernels, respectively. Smaller convolution kernels (e.g., 3$\times$3) excel at detecting fine-grained details in medical images, such as subtle lesion areas, while larger kernels (e.g., 9$\times$9) are more effective at capturing broader structures, including the overall contours of organ.

Subsequently, the features from different scales are integrated with the original input features through residual connections, followed by a ReLU activation function. This process facilitates richer multi-scale feature representations and can be formulated as follows:
\begin{equation}
    \bar{\bm{X}}_{i} = \mathrm{ReLU}(f_{k_i \times k_i}^{dwc}(\bm{X}_i) + \bm{X}_i),
\end{equation}
where $i \in \{1, 2, 3, 4\}$, $k_i = 2i + 1$ and the operation $f_{k_i \times k_i}^{dwc}$ represents the depth-wise convolution with the kernel size of $k_i$.

\subsubsection{Linear Attention Computation}
To further enhance the model's ability to locate regions of interest and suppress irrelevant information, we employ Efficient Attention \cite{shen2021efficient} to capture contextual information across the multi-scale features in each branch, respectively.

First, we reshape $\bar{\bm{X}}_{i} \in \mathbb{R}^{\sqrt{N} \times \sqrt{N} \times \frac{C}{4}}$ into $\bar{\bm{X}}_{i}^{r} \in \mathbb{R}^{N \times \frac{C}{4}}$. To extract global representations, we apply Efficient Attention to the multi-scale tokens $\bar{\bm{X}}_{i}^{r}$. Specifically, for $i$-th branch and $h$-th head, we derive the query, key, value tensor, $\bm{Q}_{i,h},\bm{K}_{i,h},\bm{V}_{i,h} \in \mathbb{R}^{N \times d}$, using specific linear projections:
\begin{equation}
    \bm{Q}_{i,h}=\bar{\bm{X}}_{i}^{r} \bm{W}_{i,h}^q,
    \bm{K}_{i,h}=\bar{\bm{X}}_{i}^{r} \bm{W}_{i,h}^k,
    \bm{V}_{i,h}=\bar{\bm{X}}_{i}^{r} \bm{W}_{i,h}^v,
\end{equation}
where  $\bm{W}_{i,h}^q, \bm{W}_{i,h}^k, \bm{W}_{i,h}^v \in \mathbb{R}^{\frac{C}{4} \times d}$ are projection weights for the query, key, value, respectively.

Next, we perform attention computation separately for each head using Efficient Attention. Formally,
\begin{equation}
    \begin{split}
        \bm{head}_{i,h} = \text{EfficientAtt}_{i,h} (\bm{Q}_{i,h}, \bm{K}_{i,h}, \bm{V}_{i,h}) ,\\
        \bm{O}_i = \text{Concat}(\bm{head}_{i,0},  \bm{head}_{i,1}, \ldots, \bm{head}_{i,h-1}) \bm{W}_i^O,
    \end{split}
\end{equation}
where $\bm{head}_{i,h}$ is the output of the $h$-th attention head in $i$-th branch. An additional linear transformation with weight matrix $\bm{W}_i^O\in \mathbb{R}^{\frac{C}{4} \times \frac{C}{4}}$ is applied to combine the outputs from all heads.

Finally, $\bm{O}_i \in \mathbb{R}^{N \times \frac{C}{4}}$ is reshaped  to image representations $\bm{O}_{i}^{r} \in \mathbb{R}^{\sqrt{N} \times \sqrt{N} \times \frac{C}{4}}$ on spatial dimension. 
This transformation facilitates subsequent convolution operations,
thereby enhancing the fusion of multi-scale features. The fusion process can be described as:
\begin{equation}
    \bm{O}^{f} = f_{1 \times 1}([w_1 \bm{O}_1^r, w_2 \bm{O}_2^r, w_3 \bm{O}_3^r, w_4 \bm{O}_4^r]),
\end{equation}
where $w_i$ are learnable weight parameters, $\left[ \cdot \right]$ represents the channel-wise concatenation, and $f_{1 \times 1}$ denotes a $1 \times1$ convolution. We then reshape the fused feature map $\bm{O}^{f} \in \mathbb{R}^{\sqrt{N} \times \sqrt{N} \times C}$ into $\bm{O} \in \mathbb{R}^{ N \times C}$ to obtain the final output tokens.

\subsection{Encoder}
The encoder of our model consists of four stages, as shown in Fig. \ref{fig4:mslanet}. The first two stages comprise Patch Embedding layers and Local Feature Extraction (LFE) blocks, while the last two stages incorporate Patch Embedding layers and Global Feature Extraction (GFE) blocks. Below, we provide a detailed description of the LFE block, the GFE block, and the Patch Embedding layer. These components are progressively stacked in a hierarchical manner, forming a four-stage pyramid structure with the configuration of [4,8,11,5]. This hierarchical stacking approach ensures that both local details and global context are effectively captured, leading to more robust and comprehensive feature representations.

\begin{figure}[htbp]
\centering
\includegraphics[width=1.0\linewidth, keepaspectratio]{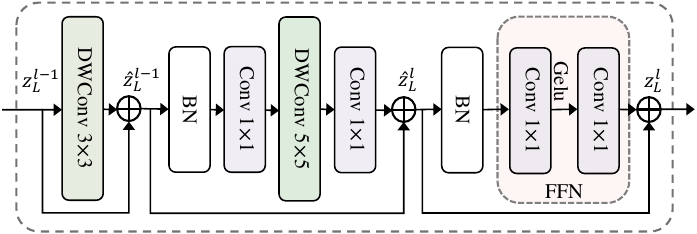}
\caption{Details of the LFE block. An LFE block consists of three key modules: a 3$\times$3 depth-wise convolution, three consecutive convolutional layers, and an FFN.}
\label{fig2:lfe}
\end{figure}

Following the Local UniFormer block \cite{li2023uniformer}, we introduce the LFE block to better leverage the local feature extraction capabilities of CNNs. The LFE block consists of three components: a 3$\times$3 depth-wise convolution, three consecutive convolutional layers, and a Feed-Forward Network (FFN), as illustrated in Fig. \ref{fig2:lfe}. The 3$\times$3 depth-wise convolution encodes relative position information, while the three consecutive convolutional layers learn local representations. Regarding the FFN, the channel dimension of its input is first expanded by a ratio of 4 through a 1$\times$1 convolution and then restored to the original dimension using another 1$\times$1 convolution. The LFE block can be formulated as:
\begin{equation}
{\hat {\bm{z}}_{L}^{l - 1}} = f_{3 \times 3}^{dwc}({\bm{z}_{L}^{l - 1}}) + {\bm{z}_{L}^{l - 1}},
\end{equation}
\begin{equation}
{\hat {\bm{z}}_{L}^l} = f_{1 \times 1}(f_{5 \times 5}^{dwc}(f_{1 \times 1}(\operatorname{BN}({\hat {\bm{z}}_{L}^{l - 1}})))) + {\hat {\bm{z}}_{L}^{l - 1}},
\end{equation}
\begin{equation}
{\bm{z}_{L}^l} = \operatorname{FFN}(\operatorname{BN}({\hat {\bm{z}}_{L}^l})) + {\hat {\bm{z}}_{L}^l},
\end{equation}
where ${\hat {\bm{z}}_{L}^{l - 1}}$, ${\hat {\bm{z}}_{L}^l}$ and ${\bm{z}_{L}^l}$ represent the outputs of the depth-wise convolution, three consecutive convolutional layers and FFN module of the $l$-{th} LFE block, respectively.

\begin{figure}[b]
\centering
\includegraphics[width=1.0\linewidth, keepaspectratio]{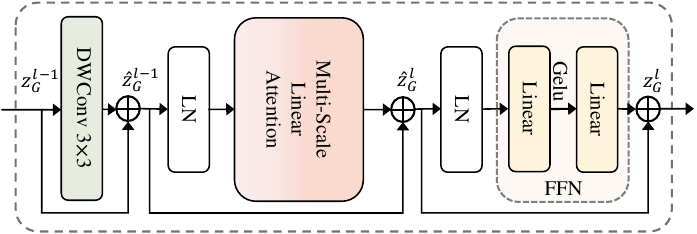}
\caption{Details of the GFE Block. A GFE block comprises three main components: a 3$\times$3 depth-wise convolution, an MSLA module, and an FFN.}
\label{fig3:gfe}
\end{figure}

Based on the MSLA module, we design the GFE blocks to learn long-range dependencies in the deeper layers. Specifically, we employ a 3$\times$3 depth-wise convolution at the beginning as relative position encoding. Subsequently, we sequentially apply an MSLA module and an FFN module with an expansion ratio of $e=4$, as shown in Fig. \ref{fig3:gfe}. The GFE block can be expressed as:
\begin{equation}
{\hat {\bm{z}}_{G}^{l - 1}} = f_{3 \times 3}^{dwc}({\bm{z}_{G}^{l - 1}}) + {\bm{z}_{G}^{l - 1}},
\end{equation}
\begin{equation}
{\hat {\bm{z}}_{G}^l} = \operatorname{MSLA}(\operatorname{LN}({\hat {\bm{z}}_{G}^{l - 1}})) + {\hat {\bm{z}}_{G}^{l - 1}},
\end{equation}
\begin{equation}
{\bm{z}_{G}^l} = \operatorname{FFN}(\operatorname{LN}({\hat {\bm{z}}_{G}^l})) + {\hat {\bm{z}}_{G}^l},
\end{equation}
where ${\hat {\bm{z}}_{G}^{l - 1}}$, ${\hat {\bm{z}}_{G}^l}$ and ${\bm{z}_{G}^l}$ represent the outputs of the depth-wise convolution, MSLA module and FFN module of the $l$-th GFE block, respectively.

Regarding the Patch Embedding layers, we apply a 4$\times$4 convolution with a stride of 4 in the first stage and a 2$\times$2 convolution with a stride of 2 in the subsequent stages. An extra Layer Normalization (LN) is added after each down-sampling convolution.

\subsection{Decoder}
\begin{figure*}[htbp]
\centering
\includegraphics[width=1.0\linewidth, keepaspectratio]{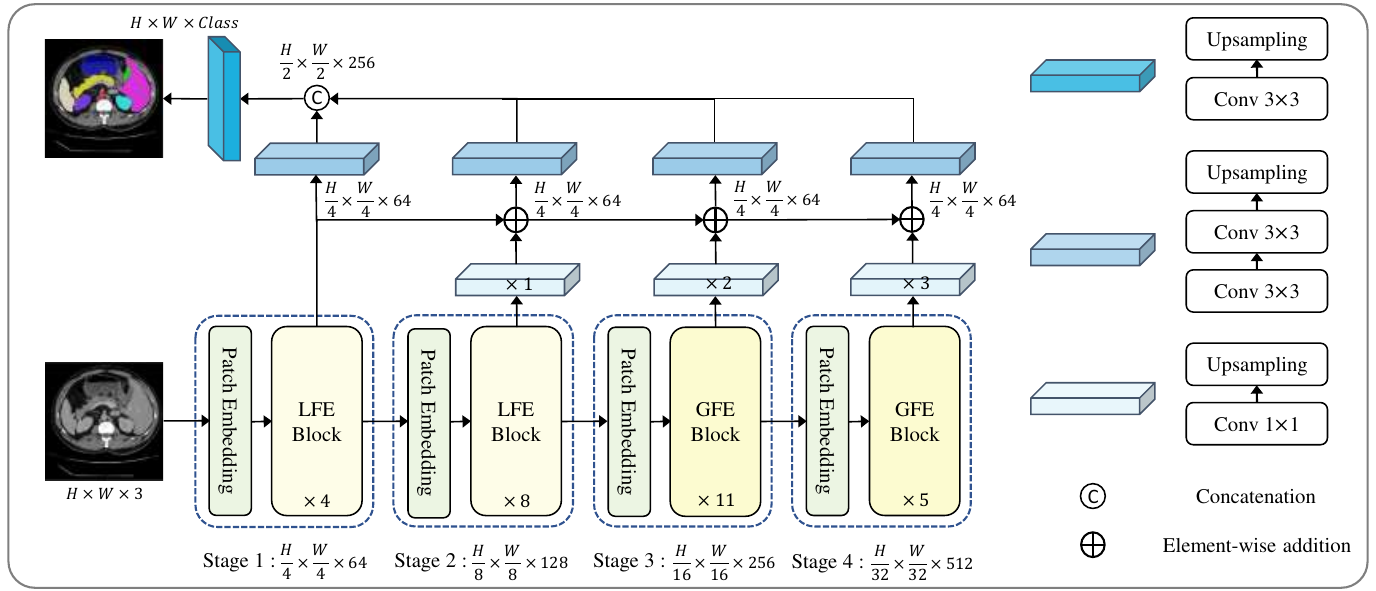}
\caption{The proposed MSLAU-Net adopts an encoder-decoder structure. The encoder integrates CNN and transformer components, utilizing LFE and GFE blocks for local and global feature extraction, respectively. The decoder employs a top-down aggregation mechanism to aggregate multi-level features from the corresponding stages of the encoder. These features are then upsampled to the original image resolution, producing the final mask prediction.}
\label{fig4:mslanet}
\end{figure*}

The decoder consists of convolution and bilinear upsampling operations, designed to perform multi-level feature aggregation and restore the resolution using a top-down feature aggregation mechanism, as shown in Fig. \ref{fig4:mslanet}. To effectively aggregate multi-level features, we align the outputs from the second to the fourth stages of the encoder with the output of the first stage in terms of both channel dimensions and spatial resolutions. Specifically, we apply one, two, and three convolution blocks to the outputs from the second, third, and fourth stages, respectively. Each block consists of a 1$\times$1 convolution layer followed by a bilinear upsampling layer. The 1$\times$1 convolution layer reduces the channel dimension by half, while the bilinear upsampling layer performs 2$\times$ upsampling. This alignment is crucial for maintaining uniform feature representations across all stages, facilitating more effective multi-scale feature fusion, and improving overall model performance.
After obtaining uniform channel dimensions and spatial resolutions, we sequentially perform a top-down aggregation operation on the outputs (i.e., feature maps) of the second to fourth stages using element-wise addition, thereby enhancing interactions across different stages. 

Next, all aggregated feature maps with uniform channel dimensions and spatial resolutions are fed into another convolution block, which consists of two successive 3$\times$3 convolution layers and a bilinear upsampling layer. This configuration increases the spatial resolution by 2$\times$ while keeping the channel dimension unchanged. The resulting outputs from each branch are then concatenated along the channel dimension, increasing the channel dimension by 4$\times$.
Subsequently, the concatenated feature maps are then fed into the third convolution block, which consists of a 3$\times$3 convolution layer and a bilinear upsampling layer, to recover the full resolution $H \times W$ for predicting the final segmentation outcome.

\subsection{Architecture Overview}
Our architecture, named MSLAU-Net, is a hybrid CNN-Transformer model that consists of the encoder and decoder described above. MSLAU-Net seamlessly integrates multi-scale feature extraction and multi-level feature aggregation within an encoder-decoder framework. The encoder adopts a hierarchical pyramid structure, where the first two stages consist of 4 and 8 LFE blocks, respectively, while the last two stages include 11 and 5 GFE blocks. This design ensures that our network can effectively capture both low-level local features and high-level semantic information. The decoder employs a top-down feature aggregation mechanism to integrate multi-level features from the encoder, enhancing interactions across different stages and contributing to richer feature representations. Additionally, bilinear upsampling and channel-wise concatenation are utilized to increase both the spatial resolution and channel dimensions.  The entire network is carefully designed to leverage the strengths of both CNNs and transformers, ensuring efficient multi-scale feature extraction and robust multi-level feature aggregation while maintaining low computational complexity.

\subsection{Loss Function}
We utilize different loss functions for each dataset in our experiments. For the CVC-ClinicDB dataset, we optimize MSLAU-Net using only the dice loss ($\cal L$${_{dice}}$). It can be defined as follows:
 \begin{equation}
 {\mathcal{L}_{dice}} = 1 - \sum\limits_k^K {\frac{{2{\omega _k}\sum\nolimits_i^N {p(k,i)g(k,i)} }}{{\sum\nolimits_i^N {{p^2}(k,i) + \sum\nolimits_i^N {{g^2}(k,i)} } }}},
 \end{equation}
where $N$ is the number of pixels, $g(k,i) \in (0,1)$ indicates the ground truth label, and $p(k,i) \in (0,1)$ represents the predicted probability for class. $K$ is the number of class,  and $\sum\nolimits_k {{\omega _k}}$ = 1 is weight sum of all classes.

For the Synapse and ACDC datasets, we adopt a hybrid loss function that integrates both dice loss ($\cal L$${_{dice}}$) and cross-entropy ($\cal L$${_{ce}}$) loss to effectively address the issue of class imbalance. The cross-entropy loss is given by:
\begin{equation}
\begin{aligned}
\mathcal{L}_{ce} = & -\frac{1}{N}\sum_{i=1}^{N}(g(k,i) \cdot \log(p(k,i)) \\
& + (1 - g(k,i)) \cdot \log(1 - p(k,i))).
\end{aligned}
\end{equation}
The overall hybrid loss function used for training on the Synapse dataset is then formulated as:
\begin{equation}
\mathcal{L} = \lambda {\mathcal{L}_{dice}} + (1-\lambda) {\mathcal{L}_{ce}},
\end{equation}
where $\lambda$ is a weighted factor that balances the impact of $\mathcal{L}_{dice}$ and $\mathcal{L}_{ce}$. In our all experiments, the ${\omega _k}$ and $\lambda$ are empirically set as $\frac{1}{K}$ and 0.6, respectively.

\section{Experiments}
\label{sec:experiments}
\subsection{Datasets}
\textbf{Synapse Multi-Organ Segmentation Dataset}: The dataset is sourced from the MICCAI 2015 Multi-Atlas Abdomen Labeling Challenge and consists of 30 abdominal CT scans, totaling 3,779 axial slices. Each scan has a voxel size of ([0.54--0.54]$\times$[0.98--0.98]$\times$[2.5--5.0]) ${\operatorname{mm}^3}$ and contains between 85 and 198 slices, each with a resolution of 512$\times$512 pixels. Following \cite{cao2022swin,chen2021transunet}, the dataset is split into 18 cases (2,212 slices) for training and 12 cases for testing. We evaluate our method by reporting the average Dice-Similarity Coefficient (DSC) and average Hausdorff Distance (HD) across eight abdominal organs: the aorta, gallbladder, spleen, left kidney, right kidney, liver, pancreas, and stomach.

\textbf{Automated Cardiac Diagnosis Challenge Dataset}: 
The ACDC dataset consists of MRI scans acquired from 100 patients with diverse pathologies. Each scan is manually annotated for three critical regions: left ventricle (LV), right ventricle (RV), and myocardium (Myo). The dataset is partitioned into 70 training samples, 10 validation samples, and 20 testing samples. Following \cite{cao2022swin,chen2021transunet}, we evaluate the performance of our method on these three cardiac structures using the average Dice Similarity Coefficient (DSC) as the evaluation metric.

\textbf{CVC-ClinicDB Dataset}: The dataset served as the official training dataset for the MICCAI 2015 Sub-Challenge on Automatic Polyp Detection. It comprises 612 images extracted from colonoscopy videos, which are randomly partitioned into three subsets: 490 images for training, 61 for validation, and 61 for testing. Each image in the dataset is accompanied by a ground truth mask that delineates the regions of polyps. On this dataset, we adopt the metrics include Mean Intersection over Union (mIoU), DSC, Accuracy, Precision, and Recall.

\subsection{Implementation Details}
The MSLAU-Net is implemented using Python 3.10 and PyTorch 2.0. We train MSLAU-Net and its various ablation variants on an NVIDIA GeForce RTX 3090 GPU with 24 GB of memory. During the training process, the model weights are initialized using pre-trained weights from ImageNet \cite{deng2009imagenet}. For the proposed attention mechanism, its multi-scale convolution kernels are configured as [3, 5, 7, 9] based on the results of the ablation study. For the Synapse multi-organ segmentation dataset, the input images are resized to 224$\times$224. The model is trained for 400 epochs using Stochastic Gradient Descent (SGD) optimizer with a batch size of 24, an initial learning rate of 0.05, a momentum of 0.9, and a weight decay of $1e-4$. This optimizer choice aligns with common practices for the Synapse multi-organ segmentation benchmark \cite{chen2021transunet,cao2022swin}, where SGD has shown better generalization across diverse organ structures.
Similarly, for the ACDC dataset, the input images are resized to 224$\times$224, but we employ the AdamW optimizer \cite{loshchilov2017decoupled} with a batch size of 24, setting the initial learning rate to $3e-4$ and the weight decay to $5e-4$, training the model for 400 epochs. For the cardiac segmentation task with moderate data size and fine-grained anatomical details, AdamW is adopted for its adaptive learning rates that contribute to stabilized training and improved convergence.
Additionally, for the CVC-ClinicDB dataset, the input images are resized to 256$\times$256. To enhance data diversity, we apply several data augmentation techniques with a probability of 0.25, including horizontal flipping, vertical flipping, rotation, and cutout. The model is then trained for 200 epochs using the AdamW optimizer with a batch size of 8, a weight decay of $5e-4$, and an initial learning rate of $1e-4$. AdamW is selected for its demonstrated efficiency in optimization scenarios with limited training data, ensuring effective convergence for the polyp segmentation task.

\subsection{Comparison on Synapse Multi-Organ Segmentation Dataset}
The segmentation performance of the proposed MSLAU-Net is quantitatively evaluated on the Synapse Multi-Organ Segmentation dataset and compared to several state-of-the-art approaches.
These involve CNN-based approaches such as U-Net \cite{ronneberger2015u} and Att-UNet \cite{oktay2018attention}, Mamba-based approaches like VM-UNet \cite{ruan2024vm}, Transformer-based methods like Swin-Unet \cite{cao2022swin} and MISSFormer \cite{huang2021missformer}, and hybrid CNN-Transformer approaches including TransUNet \cite{chen2021transunet}, HiFormer \cite{heidari2023hiformer}, PVT-CASCADE \cite{rahman2023medical}, and BRAU-Net++ \cite{lan2024brau}. 
The experimental results are presented in Table \ref{tab1:synapse}. As shown in Table \ref{tab1:synapse}, our method achieves the highest DSC of 83.18\%, significantly outperforming all other widely adopted techniques. This demonstrates the effectiveness of MSLAU-Net in modeling both local features and global dependencies.
In terms of HD, MSLAU-Net also performs admirably, achieving a value of 17.00 mm. While this is slightly higher than the best-performing HiFormer (14.70 mm), it still reflects strong boundary localization capabilities. 
Furthermore, MSLAU-Net exhibits promising performance across multiple organs. Notably, it achieves the highest DSC for the liver (94.82\%) and the right kidney (84.67\%), underscoring its effectiveness in handling complex anatomical structures. 
Additionally, MSLAU-Net is highly parameter-efficient, with only 21.90 M parameters, making it a practical and appealing choice for real-world applications. 
These results show that MSLAU-Net not only achieves state-of-the-art performance but also maintains computational efficiency, positioning it as a promising solution for multi-organ segmentation tasks.

\begin{figure*}[htbp]
\centering
\includegraphics[width=1.0\linewidth, keepaspectratio]{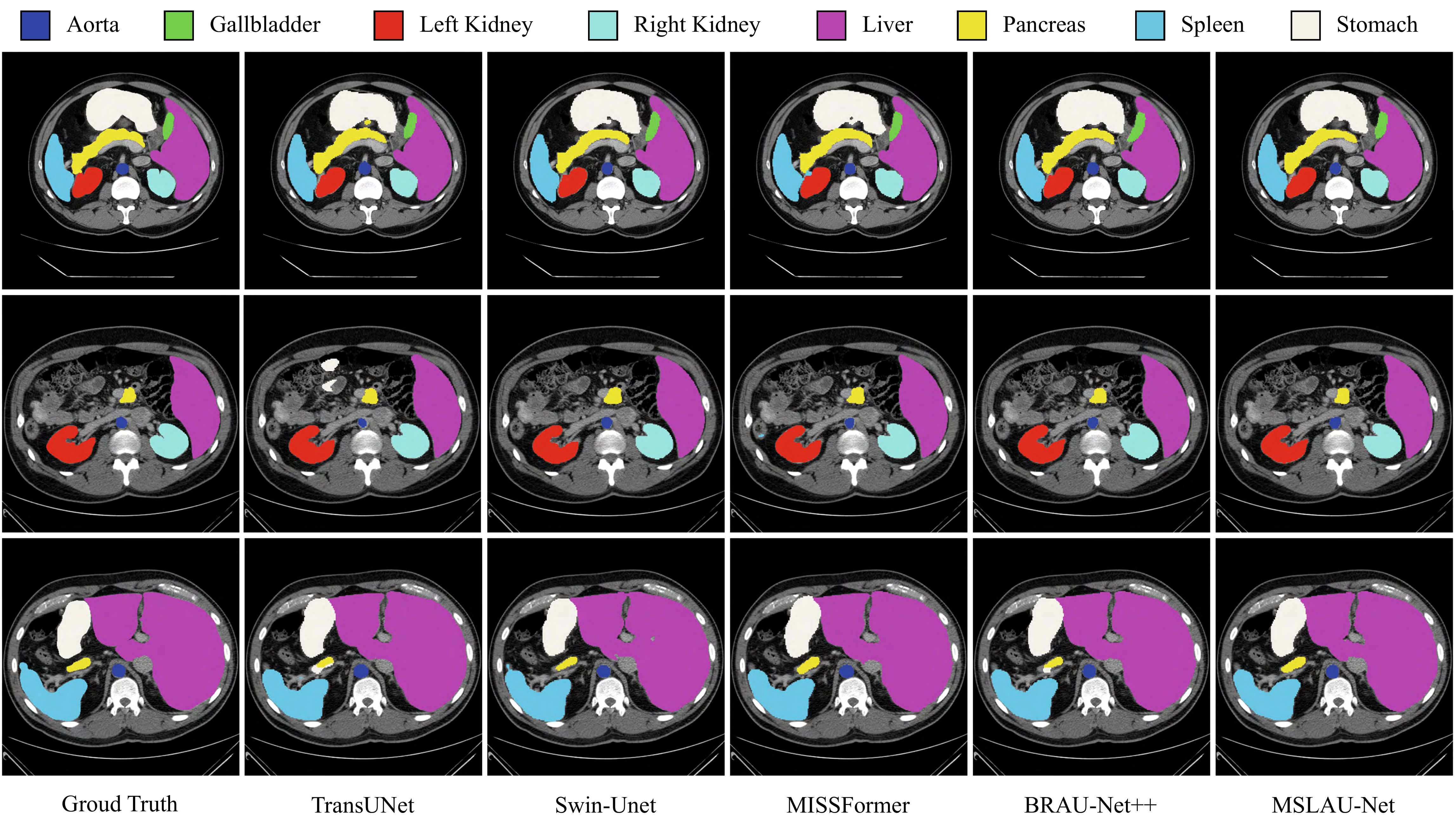}
\caption{Qualitative results of different methods on Synapse dataset. Our MSLAU-Net captures organ boundaries more accurately and demonstrates superior detail-handling capabilities. Best viewed in color with zoom-in.}
\label{fig5:synapse_results}
\end{figure*}

Qualitative results of different methods on the Synapse dataset are shown in Fig. \ref{fig5:synapse_results}. One can see that our method generates more accurate segmentation maps for organs such as the gallbladder, right kidney, liver, and pancreas. These results demonstrate that MSLA performs well in capturing the features of both small and large targets.
Moreover, MSLAU-Net effectively learns both local details and global semantic information, thereby yielding better segmentation results. Specifically, the proposed approach not only delineates the boundaries of organs more accurately but also maintains consistency across the entire segmentation map. These qualities highlight the effectiveness of the MSLA mechanism in enhancing the model's ability to handle complex anatomical structures.

\begin{table*}[htbp]
\centering
\caption{Quantitative results of our approach against other state-of-the-art methods on Synapse dataset. The symbol $\uparrow$ indicates the larger the better. The symbol $\downarrow$ \iffalse indicates \fi denotes the smaller the better. The best result is in \textbf{blod}, and the second best is \underline{underlined}.}
\resizebox{1.0\linewidth}{!}{%
\begin{tabular}{l|c|cc|cccccccc}
\toprule
Methods & Params (M) & DSC (\%) $\uparrow$  & HD (mm) $\downarrow$  & Aorta &Gallbladder & Kidney (L)  & Kidney (R)  & Liver  & Pancreas & Spleen  & Stomach\\
\midrule
U-Net \cite{ronneberger2015u} &14.80   &76.85   &39.70 &\underline{89.07} &69.72 &77.77 &68.60 &93.43 &53.98 &86.67 &75.58 \\
Att-UNet \cite{oktay2018attention} &34.88  &77.77  &36.02 &\textbf{89.55} &68.88 &77.98 &71.11 &93.57 &58.04 &87.30 &75.75 \\
VM-UNet \cite{ruan2024vm} &27.43 &81.08 &19.21 &86.40 &69.41 &\underline{86.16} &\underline{82.76} &94.17 &58.80 &89.51 &81.40 \\
TransUNet \cite{chen2021transunet} &105.28 &77.48   &31.69 &87.23 &63.13 &81.87 &77.02 &94.08 &55.86 &85.08 &75.62 \\
Swin-Unet \cite{cao2022swin} &27.17 &79.13  &21.55 & 85.47 &66.53  & 83.28 &79.61  &94.29 & 56.58 &90.66  & 76.60 \\
HiFormer \cite{heidari2023hiformer} &25.51 &80.39  &\textbf{14.70} & 86.21 &65.69  & 85.23 &79.77  &94.61 & 59.52 &90.99 & 81.08 \\
PVT-CASCADE \cite{rahman2023medical} &35.28 &81.06 &20.23& 83.01 &\underline{70.59} & 82.23 &80.37  &94.08 & 64.43 &90.10  & \textbf{83.69} \\
MISSFormer \cite{huang2021missformer} &42.46 &81.96  &18.20& 86.99 &68.65  &85.21  &82.00  &94.41 & \underline{65.67} &\textbf{91.92} & 80.81 \\
BRAU-Net++ \cite{lan2024brau} &62.63 &\underline{82.47}  &19.07 & 87.95 &69.10  &\textbf{87.13} &81.53 &\underline{94.71}  &65.17 &\underline{91.89} &\underline{82.26}\\
\midrule
MSLAU-Net (Ours) &21.90 &\textbf{83.18} & \underline{17.00} &88.68 &\textbf{73.95} &85.54 &\textbf{84.57} &\textbf{94.82} &\textbf{65.79} &91.20 &80.87 \\
\bottomrule
\end{tabular} }
\label{tab1:synapse}
\end{table*}

\subsection{Comparison on Automated Cardiac Diagnosis Challenge Dataset}
The comparisons of our proposed method with previous state-of-the-art methods on the ACDC dataset are given in Table \ref{tab2:acdc}. The results show that MSLAU-Net achieves an overall DSC of 92.13\%, surpassing all other competing methods. Notably, MSLAU-Net exhibits the best performance in myocardial (Myo) and left ventricle (LV) segmentation, achieving a DSC of 90.06\% and 95.95\%, highlighting its distinct advantage in handling specific regions. These findings indicate that MSLAU-Net is a highly effective and robust model for cardiac image segmentation tasks.

\begin{table}[htbp]
\centering
\caption{Quantitative results of different methods on ACDC dataset.} 
\resizebox{0.8\linewidth}{!}{
\begin{tabular}{l|c|ccc}
\toprule
Methods & DSC (\%) $\uparrow$ & RV & Myo &LV 
\\
\midrule
R50+U-Net \cite{chen2021transunet}   &87.55  &87.10 &80.63 &94.92
\\
R50+Att-UNet \cite{chen2021transunet}  &86.75  &87.58 &79.20 &93.47
\\
VM-UNet \cite{ruan2024vm} &90.42 &88.93 &87.35 &94.98
\\
TransUNet \cite{chen2021transunet} &89.71  &88.86 &84.54 &95.73
\\
Swin-Unet \cite{cao2022swin}  &90.00 &88.55 &85.62 &95.83
\\
MISSFormer \cite{huang2021missformer}  &90.86 &89.55 &88.04 &94.99
\\
PVT-CASCADE \cite{rahman2023medical}  &91.46 &88.90 &\underline{89.97} &95.50
\\
BRAU-Net++ \cite{lan2024brau}  &\underline{92.07} &\textbf{90.72} &89.57 &\underline{95.90}
\\
\midrule
MSLAU-Net (Ours) &\textbf{92.13} &\underline{90.38}  &\textbf{90.06} &\textbf{95.95}   \\
\bottomrule
\end{tabular}}
\\
\label{tab2:acdc}
\end{table}

\subsection{Comparison on CVC-ClinicDB Dataset}
\begin{figure*}[htbp]
\centering
\includegraphics[width=1.0\linewidth, keepaspectratio]{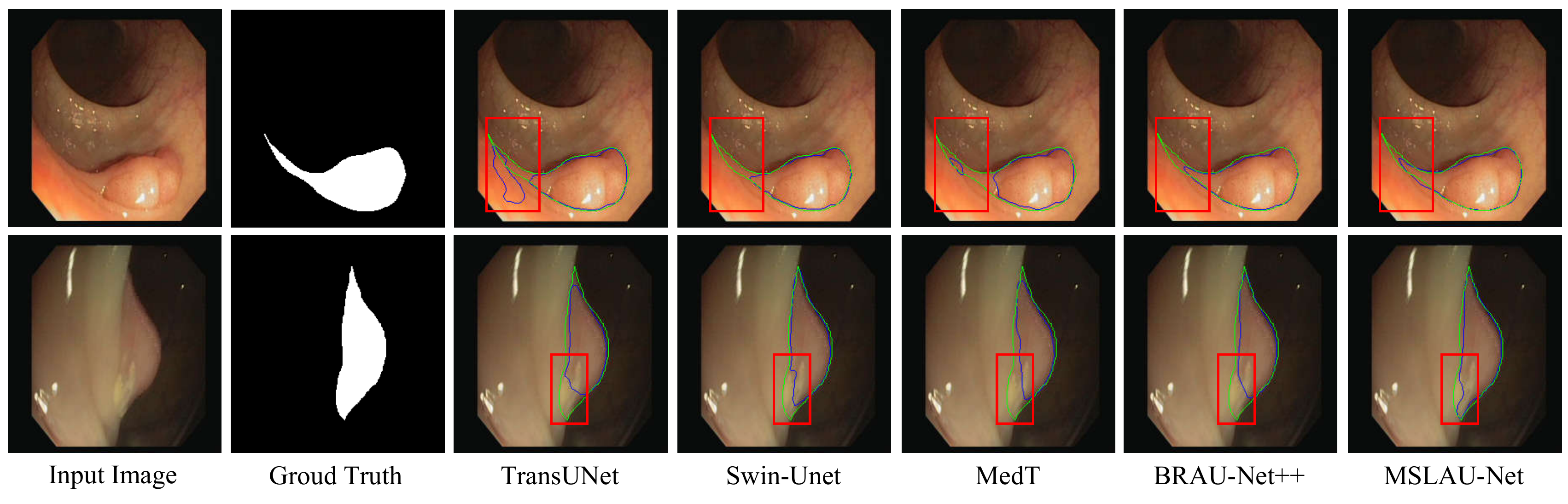}
\caption{Qualitative results of different methods on CVC-ClinicDB dataset. Ground truth boundaries are shown in \textcolor{green}{green}, and predicted boundaries are shown in \textcolor{blue}{blue}. Our MSLAU-Net demonstrates better performance in matching the ground truth compared to other state-of-the-art methods. Best viewed in color with zoom-in.}
\label{fig6:cvc_results}
\end{figure*}

According to the comparative experimental results in Table \ref{tab3:CVC}, MSLAU-Net exhibits outstanding segmentation performance on the CVC-ClinicDB dataset. The method achieves a mIoU of 88.68 and a DSC of 93.03\%, while also excelling in other key metrics such as accuracy, precision, and recall. Furthermore, as illustrated in Fig. \ref{fig6:cvc_results}, the segmentation results produced by MSLAU-Net closely align with the ground truth. These results confirm the robust capability and high reliability of MSLAU-Net in handling complex medical image segmentation tasks.

\begin{table}[htbp]
\centering
\caption{Quantitative results of different methods on CVC-ClinicDB dataset.} 
\resizebox{1.0\linewidth}{!}{
\begin{tabular}{l|ccccc}
\toprule
Methods & mIoU $\uparrow$ & DSC $\uparrow$ &Accuracy $\uparrow$& Precision $\uparrow$ &Recall $\uparrow$ 
\\
\midrule
U-Net \cite{ronneberger2015u}   &80.91  &87.22&98.45 &88.24 &89.35
\\
Att-UNet \cite{oktay2018attention}  &83.54  &89.57 &98.64 &90.47 &90.10
\\
MedT \cite{valanarasu2021medical}  &81.47  &86.97 &98.44 &89.35 &90.04
\\
VM-UNet \cite{ruan2024vm} &86.24 &90.88 &98.78 &\underline{91.86} &91.45
\\
TransUNet \cite{chen2021transunet} &79.95  &86.70 &98.25 &87.63 &87.34
\\
Swin-Unet \cite{cao2022swin}  &84.85 &88.21 &98.72 &90.52 &91.13
\\
BRAU-Net++ \cite{lan2024brau} &\underline{88.17} &\underline{92.94} &\underline{98.83} &\textbf{93.84} &\textbf{93.06}
\\
\midrule
MSLAU-Net (Ours) &\textbf{88.68} &\textbf{93.03}  &\textbf{98.84} &\textbf{93.84} &\underline{92.65}  \\
\bottomrule
\end{tabular}}
\label{tab3:CVC}
\end{table}

\subsection{Ablation Study}
In this section, we conduct extensive ablation studies on the Synapse dataset to thoroughly evaluate each component of MSLAU-Net. Our analysis encompasses the effectiveness of pre-trained weights, comparative performance of our MSLA module against other linear attention mechanisms, structural designs within the encoder, decoder architecture choices, multi-scale combinations in MSLA, and model scales. All experiments, except for the pre-trained weights analysis, are conducted from scratch.

\subsubsection{Effectiveness of Pre-trained Weights}
To evaluate the specific effectiveness of pre-trained weights on improving model performance, we conduct an ablation study on the Synapse dataset. As shown in Table \ref{tab4:ablation_pre}, when no pre-trained weights are used (i.e., random initialization), the method have to learn all parameters from scratch, resulting in inferior performance compared to the case where pre-trained weights are utilized. In contrast, initializing the model with weights pre-trained on ImageNet significantly enhanced its performance. Specifically, the DSC is improved by approximately 3.77\%, and the HD is reduced by 5.88 mm. Due to resource limitations, the following ablation studies are conducted without using pre-trained weights.

\begin{table*}[htbp]
\centering
\caption{Ablation study of pre-trained weights on Synapse dataset.}
\resizebox{1.0\linewidth}{!}{
\begin{tabular}{c|c|cc|cccccccc}
\toprule
Model Scale & Pre-trained  & DSC (\%) $\uparrow$  & HD (mm) $\downarrow$  & Aorta &Gallbladder & Kidney (L)  & Kidney (R)  & Liver  & Pancreas & Spleen  & Stomach 
\\
\midrule
\multirow{2}{*}{Base} &  w/o  &79.41 &22.88 &87.16 &70.26 &82.06 &79.11 &94.08 &55.73 &90.88 &76.02 
\\
~  & w  &\textbf{83.18} & \textbf{17.00} &\textbf{88.68} &\textbf{73.95} &\textbf{85.54} &\textbf{84.57} &\textbf{94.82} &\textbf{65.79} &\textbf{91.26} &\textbf{80.87} 
\\
\bottomrule
\end{tabular}}
\label{tab4:ablation_pre}
\end{table*}

\subsubsection{Effectiveness of Structure Designs within Encoder}
We conduct an ablation study to evaluate the effectiveness of different configurations of LFE and GFE blocks stage by stage within the encoder. The results are presented in Table \ref{tab5:ablation_encoder}. One can see that using only LFE blocks (i.e., the configuration LLLL) or only GFE blocks (i.e., the configuration GGGG) for all stages yields relatively low performance. The best segmentation performance is achieved with the configuration LLGG.
The phenomena reveal that while transformers excel at global feature modeling, they are relatively less effective than CNNs in extracting local representations, potentially leading to redundant attention. As expected, the optimal performance is achieved by leveraging CNNs in the first two stages and transformers in the last two stages. This configuration effectively balances local detail extraction and global context understanding by leveraging the strengths of both architectures.

\begin{table}[htbp]
\centering
\caption{Ablation study of encoder structure designs on Synapse dataset. ``L'' represents the LFE block, ``G'' indicates the GFE block, and ``S'' denotes Stage.}
\resizebox{1.0\linewidth}{!}{
\begin{tabular}{cccc|cc|cc}
\toprule
\multicolumn{4}{c|}{Types} & \multirow{2}{*}[-0.6ex]{Params (M)} & \multirow{2}{*}[-0.6ex]{FLOPs (G)} &\multirow{2}{*}[-0.6ex]{DSC (\%) $\uparrow$}  & \multirow{2}{*}[-0.6ex]{HD (mm) $\downarrow$} 
\\
\cmidrule(lr){1-4}
S1 & S2 & S3 & S4 & ~ & ~  & ~ & ~ \\
\midrule
L & L & L & L  &23.34  &5.02 &77.34 &27.16
\\
L & L &L & G  &22.39  &5.02 &78.55 &24.94
\\
L& L & G & G  &21.90  &5.05 &\textbf{79.41} &\textbf{22.88}
\\
L & G & G &G  &21.81  &5.10 &\underline{78.67} &\underline{23.54}
\\
G & G & G & G  &21.81  &5.14 &76.80 &29.66
\\
\bottomrule
\end{tabular}}
\label{tab5:ablation_encoder}
\end{table}

\subsubsection{Effectiveness of MSLA against Other Linear Attention Mechanisms}
To evaluate the advantages of our MSLA module, we conducted a comprehensive comparison with mainstream linear attention mechanisms within our MSLAU-Net architecture on the Synapse dataset. As summarized in Table \ref{tab6:ablation_attention}, the proposed MSLA, built upon Efficient Attention, achieves the highest DSC (79.41\%) and the competitive HD (22.88 mm), while requiring the fewest parameters (21.90 M) and the lowest computational cost (5.05 GFLOPs). 

It is worth emphasizing that the base Efficient Attention \cite{shen2021efficient} mechanism itself already exhibits favorable efficiency, with fewer parameters and lower computational load than most compared methods. By incorporating parallel multi-scale branches, the proposed MSLA further enhances the capability to capture hierarchical features in medical images—ranging from fine-grained textures to organ-level semantics—while preserving the global receptive field and linear complexity. Experimental results confirm that the multi-scale design based on Efficient Attention \cite{shen2021efficient} effectively improves overall segmentation performance while maintaining high computational efficiency, making it particularly suitable for medical image segmentation tasks.

\begin{table}[htbp]
\centering
\caption{Ablation study of different linear attention mechanisms in MSLAU-Net on Synapse dataset.} 
\resizebox{1.0\linewidth}{!}{
\begin{tabular}{l|cc|cc}
\toprule
Linear Attention & Param (M) & FLOPs (G) & DSC (\%) $\uparrow$ & HD (mm) $\downarrow$ \\
\midrule
Hydra Attn \cite{bolya2022hydra} & 27.27 & 5.43 & 76.79 & 28.31 \\
Efficient Attn \cite{shen2021efficient} & 27.27 & 5.52 & 77.76 & 27.27 \\
Multi-Scale Linear Attn \cite{cai2023efficientvit} & 29.91 & 5.74 & 78.68 & \underline{21.96} \\
Focused Linear Attn \cite{han2023flatten} & 27.31 & 5.54 & \underline{79.18} & \textbf{21.72} \\
Agent Attn \cite{han2024agent} & 27.91 & 5.57 & 78.31 & 24.07 \\
\midrule
MSLA (Ours) & 21.90 & 5.05 & \textbf{79.41} & 22.88 \\
\bottomrule
\end{tabular}}
\label{tab6:ablation_attention}
\end{table}

\subsubsection{Effectiveness of Different Scale Combinations in MSLA}
We conduct a detailed ablation study to evaluate the impact of different scale combinations in MSLA on segmentation performance. Specifically, we first assess the performance without employing any multi-scale strategy, i.e., using only Efficient Attention, which achieves a DSC of 77.76\% and an HD of 27.27 mm. Subsequently, we explore dual-branch and four-branch multi-scale strategies, where convolution kernel sizes 1, 3, 5, 7, and 9 are used in different combinations. The results are shown in Table \ref{tab7:ablation_ms}. It can be seen that in the dual-branch strategy, the best performance is achieved with kernel sizes 5 and 7, resulting in a DSC of 78.32\% and an HD of 24.92 mm, while for the four-branch strategy, setting the kernel sizes to [3, 5, 7, 9] yields relatively optimal performance, achieving a DSC of 79.41\% and an HD of 22.88 mm. These results indicate that incorporating a multi-scale design, particularly the four-branch strategy, significantly enhances the performance of Efficient Attention. The improvement can be attributed to the ability of multi-scale strategies to capture features at various levels of detail.

\begin{table}[htbp]
\centering
\caption{Ablation study of different scale combinations in MSLA module on Synapse dataset.}  
\resizebox{1.0\linewidth}{!}{
\begin{tabular}{c|c|ccccc|cc}
\toprule
\multirow{2}{*}[-0.6ex]{Multi-Scale} & \multirow{2}{*}[-0.6ex]{Branches} & \multicolumn{5}{c|}{Kernel Size} &  \multirow{2}{*}[-0.6ex]{DSC (\%) $\uparrow$} &  \multirow{2}{*}[-0.6ex]{HD (mm) $\downarrow$ }\\
\cmidrule(lr){3-7}
~ & ~ & 1 & 3 & 5 & 7 & 9 & ~ & ~ \\
\midrule
w/o & - & - & - & - & - & - & 77.76 & 27.27 \\
\midrule
\multirow{6}{*}{w} & \multirow{4}{*}{2} & \ding{51} & \ding{51} &  &  &  & 77.55 & 26.35 \\
~ & ~ &  & \ding{51} & \ding{51} &  &  & 78.19 & 26.99 \\
~ & ~ &  &  & \ding{51} & \ding{51} &  & \underline{78.32} & 24.92 \\
~ & ~ &  &  &  & \ding{51} & \ding{51} & 77.65 & 24.89 \\
\cmidrule(lr){2-9}
~ & \multirow{2}{*}{4} & \ding{51} & \ding{51} & \ding{51} & \ding{51} &  & 78.21 & \textbf{21.84} \\
~ & ~ &  & \ding{51} & \ding{51} & \ding{51} & \ding{51} & \textbf{79.41} & \underline{22.88} \\
\bottomrule
\end{tabular}}
\label{tab7:ablation_ms}
\end{table}

\subsubsection{Effectiveness of Decoder Architectures}
To evaluate our decoder's effectiveness, we replaced it with a standard U-Net decoder while keeping the same encoder. As shown in Table \ref{tab8:ablation_decoder}, our decoder improves DSC by 1.29\% and reduces HD by 1.46 mm, demonstrating that the performance gains stem from the decoder design itself. Unlike U-Net's direct skip connections that concatenate encoder and decoder features at corresponding levels, our top-down aggregation employs lightweight convolutions with progressive upsampling for deeper feature fusion. This enables more effective integration of semantic information across levels, leading to superior spatial detail recovery and semantic coherence.

\begin{table}[htbp]
\centering
\caption{Ablation study of decoder architectures on Synapse dataset.} 
\resizebox{1.0\linewidth}{!}{
\begin{tabular}{l|c|c}
\toprule
Decoder Architecture & DSC (\%) $\uparrow$ & HD (mm) $\downarrow$ \\
\midrule
Standard U-Net Decoder \cite{ronneberger2015u} & 78.12 & 24.34 \\
Top-Down Aggregation Decoder (Ours) & \textbf{79.41} & \textbf{22.88} \\
\bottomrule
\end{tabular}}
\label{tab8:ablation_decoder}
\end{table}

\subsubsection{Effectiveness of Model Scales}
We use two model scales, named ``Base'' and ``Small'' models, to evaluate the impact of the model scale on segmentation performance. Their configurations and results are presented in Table \ref{tab9:ablation_model}. We can see that despite having significantly fewer parameters (14.39 M vs. 21.90 M) and lower computational complexity (3.73 GFLOPs vs. 5.05 GFLOPs), the ``Small'' variant still achieves highly competitive performance (79.41\% vs. 76.96\% on DSC and 22.88 mm vs. 25.32 mm on HD). 

For clinical deployment considerations, we evaluate inference speed on the test set, which comprises all 2D slices from 12 cases (1,567 slices in total). Each slice is processed three times at 224$\times$224 resolution, and the results are averaged. Both models achieve  desired inference speeds, with the ``Small'' variant reaching 28.54 FPS (35.04 ms/slice) and the ``Base'' model achieving 23.19 FPS (43.12 ms/slice).

However, for optimal performance, we select the ``Base'' model as our final architecture. Additionally, it is worth noting that, due to computational resource constraints, the reported performance corresponds to the model trained without using pre-trained weights.

\begin{table*}[htbp]
\centering
\caption{Ablation study of model scales on Synapse dataset.}
\resizebox{0.75\linewidth}{!}{
\begin{tabular}{c|c|cc|cc|cc}
\toprule
Model Scale & Depth & Params (M) & FLOPs (G) & Inference Time (ms) & FPS & DSC (\%) $\uparrow$ & HD (mm) $\downarrow$ \\
\midrule
Small & [3,4,8,3] & 14.39 & 3.73 & 35.04 & 28.54 & 76.96 & 25.32 \\
Base & [4,8,11,5] & 21.90 & 5.05 & 43.12 & 23.19 & \textbf{79.41} & \textbf{22.88} \\
\bottomrule
\end{tabular}}
\label{tab9:ablation_model}
\end{table*}

\subsection{Visualization}
\begin{figure}[htbp]
\centering
\includegraphics[width=1.0\linewidth, keepaspectratio]{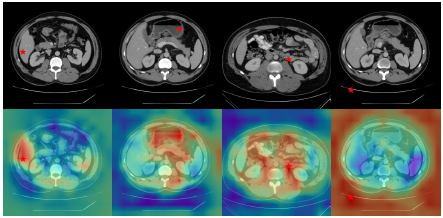}
\caption{The top row presents different slices on Synapse dataset, with \textcolor{red}{red} stars indicating the positions of the query. The bottom row displays the corresponding attention heatmaps.}
\label{fig7:attention_map}
\end{figure}

We visualize the areas of greatest concern from Stage 3 with respect to a specified query, as shown in Figure \ref{fig7:attention_map}. For queries on organ regions, marked by red stars in the first and second slices, the module accurately focuses on the corresponding regions of the organ, as indicated by high-intensity areas in the heatmap. When the query is placed in non-organ areas in the third slice, the heatmap demonstrates that the module precisely attends to these corresponding non-organ regions. Similarly, for queries located in the background in the fourth slice, the module exclusively focuses on the background area. 
These visualizations highlight the module's ability to effectively identify and concentrate on specific target regions, which is helpful to understand how our MSLA module works.

\section{Conclusions}
\label{sec:conclusion}
In this paper, we propose MSLAU-Net, a hybrid CNN-Transformer architecture designed for medical image segmentation. Specifically, we introduce Multi-Scale Linear Attention (MSLA) to capture multi-scale information and perform global attention computation while maintaining low computational complexity. Additionally, we incorporate a top-down multi-level feature aggregation mechanism in the decoder, effectively fusing high-level semantic information with low-level details to enhance segmentation accuracy. Experimental results demonstrate that our method achieves state-of-the-art performance on the Synapse, ACDC, and CVC-ClinicDB datasets, validating its effectiveness and robustness. In future work, we aim to further improve this method by designing more efficient linear attention mechanisms, thereby enhancing its computational efficiency and robustness in handling complex pathological structures.

\section*{CRediT authorship contribution statement}
\textbf{Libin Lan}: Conceptualization, Funding Acquisition, Methodology, Project Administration, Software, Writing – Review \& Editing.
\textbf{Yanxin Li}: Data Curation, Investigation, Methodology, Software, Writing – Original Draft Preparation, Visualization, Writing – Review \& Editing.
\textbf{Xiaojuan Liu}: Conceptualization, Writing – Review \& Editing, Validation.
\textbf{Juan Zhou}: Formal Analysis, Writing – Review \& Editing, Validation.
\textbf{Jianxun Zhang}: Writing – Review \& Editing, Resources.
\textbf{Nannan Huang}: Formal Analysis, Writing – Review \& Editing, Validation.
\textbf{Yudong Zhang}: Formal analysis, Writing – Review \& Editing.

\section*{Declaration of Competing Interest}
The authors declare that they have no known competing financial interests or personal relationships that could have appeared to influence the work reported in this paper.

\section*{Data Availability Statement}
All data that support the findings of this study are included within the article (and any supplementary information files)

\section*{Acknowledgments} 
This work was supported in part by the Scientific Research Foundation of Chongqing University of Technology under Grants 2021ZDZ030 and 2023ZDZ023 and in part by the Youth Project of Science and Technology Research Program of Chongqing Education Commission of China under Grants KJQN202301145 and KJQN202301162.

\balance
\bibliography{wileyNJD-AMS}

\begin{thebibliography}{10}
\providecommand{\url}[1]{\texttt{#1}}
\providecommand{\urlprefix}{URL }
\expandafter\ifx\csname urlstyle\endcsname\relax
  \providecommand{\doi}[1]{doi:\discretionary{}{}{}#1}\else
  \providecommand{\doi}{doi:\discretionary{}{}{}\begingroup \urlstyle{rm}\Url}\fi

\bibitem{alam2017alzheimer}
S.~Alam, G.-R. Kwon, and A.~D.~N. Initiative, \textit{Alzheimer disease classification using kpca, lda, and multi-kernel learning svm}, International Journal of Imaging Systems and Technology \textbf{27} (2017), no.~2, 133--143.

\bibitem{bernal2015wm}
J.~Bernal, F.~J. S{\'a}nchez, G.~Fern{\'a}ndez-Esparrach, D.~Gil, C.~Rodr{\'\i}guez, and F.~Vilari{\~n}o, \textit{Wm-dova maps for accurate polyp highlighting in colonoscopy: Validation vs. saliency maps from physicians}, Computerized medical imaging and graphics \textbf{43} (2015), 99--111.

\bibitem{bernard2018deep}
O.~Bernard et~al., \textit{Deep learning techniques for automatic mri cardiac multi-structures segmentation and diagnosis: is the problem solved?}, IEEE transactions on medical imaging \textbf{37} (2018), no.~11, 2514--2525.

\bibitem{bolya2022hydra}
D.~Bolya, C.-Y. Fu, X.~Dai, P.~Zhang, and J.~Hoffman, \textit{Hydra attention: Efficient attention with many heads}, \textit{European conference on computer vision}, Springer, 2022, 35--49.

\bibitem{cai2023efficientvit}
H.~Cai, J.~Li, M.~Hu, C.~Gan, and S.~Han, \textit{Efficientvit: Lightweight multi-scale attention for high-resolution dense prediction}, \textit{Proceedings of the IEEE/CVF international conference on computer vision}, 2023, 17302--17313.

\bibitem{cai2024pubic}
P.~Cai, L.~Jiang, Y.~Li, X.~Liu, and L.~Lan, \textit{Pubic symphysis-fetal head segmentation network using biformer attention mechanism and multipath dilated convolution}, \textit{International Conference on Multimedia Modeling}, Springer, 2024, 243--256.

\bibitem{cao2022swin}
H.~Cao, Y.~Wang, J.~Chen, D.~Jiang, X.~Zhang, Q.~Tian, and M.~Wang, \textit{Swin-unet: Unet-like pure transformer for medical image segmentation}, \textit{European conference on computer vision}, Springer, 2022, 205--218.

\bibitem{chen2021transunet}
J.~Chen et~al., \textit{Transunet: Transformers make strong encoders for medical image segmentation}, arXiv preprint arXiv:2102.04306  (2021).

\bibitem{chen2017deeplab}
L.-C. Chen, G.~Papandreou, I.~Kokkinos, K.~Murphy, and A.~L. Yuille, \textit{Deeplab: Semantic image segmentation with deep convolutional nets, atrous convolution, and fully connected crfs}, IEEE transactions on pattern analysis and machine intelligence \textbf{40} (2017), no.~4, 834--848.

\bibitem{chen2018encoder}
L.-C. Chen, Y.~Zhu, G.~Papandreou, F.~Schroff, and H.~Adam, \textit{Encoder-decoder with atrous separable convolution for semantic image segmentation}, \textit{Proceedings of the European conference on computer vision (ECCV)}, 2018, 801--818.

\bibitem{cciccek20163d}
{\"O}.~{\c{C}}i{\c{c}}ek, A.~Abdulkadir, S.~S. Lienkamp, T.~Brox, and O.~Ronneberger, \textit{3d u-net: learning dense volumetric segmentation from sparse annotation}, \textit{Medical Image Computing and Computer-Assisted Intervention--MICCAI 2016: 19th International Conference, Athens, Greece, October 17-21, 2016, Proceedings, Part II 19}, Springer, 2016, 424--432.

\bibitem{deng2009imagenet}
J.~Deng, W.~Dong, R.~Socher, L.-J. Li, K.~Li, and L.~Fei-Fei, \textit{Imagenet: A large-scale hierarchical image database}, \textit{2009 IEEE conference on computer vision and pattern recognition}, Ieee, 2009, 248--255.

\bibitem{dosovitskiy2020image}
A.~Dosovitskiy et~al., \textit{An image is worth 16x16 words: Transformers for image recognition at scale}, arXiv preprint arXiv:2010.11929  (2020).

\bibitem{du2022swinpa}
H.~Du, J.~Wang, M.~Liu, Y.~Wang, and E.~Meijering, \textit{Swinpa-net: Swin transformer-based multiscale feature pyramid aggregation network for medical image segmentation}, IEEE Transactions on Neural Networks and Learning Systems \textbf{35} (2022), no.~4, 5355--5366.

\bibitem{gao2021utnet}
Y.~Gao, M.~Zhou, and D.~N. Metaxas, \textit{Utnet: a hybrid transformer architecture for medical image segmentation}, \textit{Medical image computing and computer assisted intervention--MICCAI 2021: 24th international conference, Strasbourg, France, September 27--October 1, 2021, proceedings, Part III 24}, Springer, 2021, 61--71.

\bibitem{gu2024mamba}
A.~Gu and T.~Dao, \textit{Mamba: Linear-time sequence modeling with selective state spaces}, \textit{First Conference on Language Modeling}, 2024.

\bibitem{gu2019net}
Z.~Gu et~al., \textit{Ce-net: Context encoder network for 2d medical image segmentation}, IEEE transactions on medical imaging \textbf{38} (2019), no.~10, 2281--2292.

\bibitem{han2023flatten}
D.~Han, X.~Pan, Y.~Han, S.~Song, and G.~Huang, \textit{Flatten transformer: Vision transformer using focused linear attention}, \textit{Proceedings of the IEEE/CVF international conference on computer vision}, 2023, 5961--5971.

\bibitem{han2024agent}
D.~Han et~al., \textit{Agent attention: On the integration of softmax and linear attention}, \textit{European Conference on Computer Vision}, Springer, 2024, 124--140.

\bibitem{hatamizadeh2022unetr}
A.~Hatamizadeh et~al., \textit{Unetr: Transformers for 3d medical image segmentation}, \textit{Proceedings of the IEEE/CVF winter conference on applications of computer vision}, 2022, 574--584.

\bibitem{heidari2023hiformer}
M.~Heidari, A.~Kazerouni, M.~Soltany, R.~Azad, E.~K. Aghdam, J.~Cohen-Adad, and D.~Merhof, \textit{Hiformer: Hierarchical multi-scale representations using transformers for medical image segmentation}, \textit{Proceedings of the IEEE/CVF winter conference on applications of computer vision}, 2023, 6202--6212.

\bibitem{huang2020unet}
H.~Huang et~al., \textit{Unet 3+: A full-scale connected unet for medical image segmentation}, \textit{ICASSP 2020-2020 IEEE international conference on acoustics, speech and signal processing (ICASSP)}, IEEE, 2020, 1055--1059.

\bibitem{huang2021missformer}
X.~Huang, Z.~Deng, D.~Li, and X.~Yuan, \textit{Missformer: An effective medical image segmentation transformer}, arXiv preprint arXiv:2109.07162  (2021).

\bibitem{jiang2020hdcb}
W.~Jiang, M.~Liu, Y.~Peng, L.~Wu, and Y.~Wang, \textit{Hdcb-net: A neural network with the hybrid dilated convolution for pixel-level crack detection on concrete bridges}, IEEE Transactions on Industrial Informatics \textbf{17} (2020), no.~8, 5485--5494.

\bibitem{katharopoulos2020transformers}
A.~Katharopoulos, A.~Vyas, N.~Pappas, and F.~Fleuret, \textit{Transformers are rnns: Fast autoregressive transformers with linear attention}, \textit{International conference on machine learning}, PMLR, 2020, 5156--5165.

\bibitem{lan2024brau}
L.~Lan, P.~Cai, L.~Jiang, X.~Liu, Y.~Li, and Y.~Zhang, \textit{Brau-net++: U-shaped hybrid cnn-transformer network for medical image segmentation}, arXiv preprint arXiv:2401.00722  (2024).

\bibitem{landman2015miccai}
B.~Landman, Z.~Xu, J.~Igelsias, M.~Styner, T.~Langerak, and A.~Klein, \textit{Miccai multi-atlas labeling beyond the cranial vault--workshop and challenge}, \textit{Proc. MICCAI multi-atlas labeling beyond cranial vault—workshop challenge}, vol.~5, Munich, Germany, 2015, 12.

\bibitem{li2023uniformer}
K.~Li et~al., \textit{Uniformer: Unifying convolution and self-attention for visual recognition}, IEEE Transactions on Pattern Analysis and Machine Intelligence \textbf{45} (2023), no.~10, 12581--12600.

\bibitem{li2018h}
X.~Li, H.~Chen, X.~Qi, Q.~Dou, C.-W. Fu, and P.-A. Heng, \textit{H-denseunet: hybrid densely connected unet for liver and tumor segmentation from ct volumes}, IEEE transactions on medical imaging \textbf{37} (2018), no.~12, 2663--2674.

\bibitem{liu2021swin}
Z.~Liu et~al., \textit{Swin transformer: Hierarchical vision transformer using shifted windows}, \textit{Proceedings of the IEEE/CVF international conference on computer vision}, 2021, 10012--10022.

\bibitem{long2015fully}
J.~Long, E.~Shelhamer, and T.~Darrell, \textit{Fully convolutional networks for semantic segmentation}, \textit{Proceedings of the IEEE conference on computer vision and pattern recognition}, 2015, 3431--3440.

\bibitem{loshchilov2017decoupled}
I.~Loshchilov and F.~Hutter, \textit{Decoupled weight decay regularization}, arXiv preprint arXiv:1711.05101  (2017).

\bibitem{milletari2016v}
F.~Milletari, N.~Navab, and S.-A. Ahmadi, \textit{V-net: Fully convolutional neural networks for volumetric medical image segmentation}, \textit{2016 fourth international conference on 3D vision (3DV)}, Ieee, 2016, 565--571.

\bibitem{oktay2018attention}
O.~Oktay et~al., \textit{Attention u-net: Learning where to look for the pancreas}, arXiv preprint arXiv:1804.03999  (2018).

\bibitem{peng2020semantic}
C.~Peng and J.~Ma, \textit{Semantic segmentation using stride spatial pyramid pooling and dual attention decoder}, Pattern Recognition \textbf{107} (2020), 107498.

\bibitem{rahman2023medical}
M.~M. Rahman and R.~Marculescu, \textit{Medical image segmentation via cascaded attention decoding}, \textit{Proceedings of the IEEE/CVF winter conference on applications of computer vision}, 2023, 6222--6231.

\bibitem{rahman2024emcad}
M.~M. Rahman, M.~Munir, and R.~Marculescu, \textit{Emcad: Efficient multi-scale convolutional attention decoding for medical image segmentation}, \textit{Proceedings of the IEEE/CVF Conference on Computer Vision and Pattern Recognition}, 2024, 11769--11779.

\bibitem{ronneberger2015u}
O.~Ronneberger, P.~Fischer, and T.~Brox, \textit{U-net: Convolutional networks for biomedical image segmentation}, \textit{Medical image computing and computer-assisted intervention--MICCAI 2015: 18th international conference, Munich, Germany, October 5-9, 2015, proceedings, part III 18}, Springer, 2015, 234--241.

\bibitem{ruan2024vm}
J.~Ruan, J.~Li, and S.~Xiang, \textit{Vm-unet: Vision mamba unet for medical image segmentation}, ACM Transactions on Multimedia Computing, Communications and Applications  (2024).

\bibitem{schlemper2019attention}
J.~Schlemper, O.~Oktay, M.~Schaap, M.~Heinrich, B.~Kainz, B.~Glocker, and D.~Rueckert, \textit{Attention gated networks: Learning to leverage salient regions in medical images}, Medical image analysis \textbf{53} (2019), 197--207.

\bibitem{shen2021efficient}
Z.~Shen, M.~Zhang, H.~Zhao, S.~Yi, and H.~Li, \textit{Efficient attention: Attention with linear complexities}, \textit{Proceedings of the IEEE/CVF winter conference on applications of computer vision}, 2021, 3531--3539.

\bibitem{srivastava2021msrf}
A.~Srivastava et~al., \textit{Msrf-net: a multi-scale residual fusion network for biomedical image segmentation}, IEEE Journal of Biomedical and Health Informatics \textbf{26} (2021), no.~5, 2252--2263.

\bibitem{valanarasu2021medical}
J.~M.~J. Valanarasu, P.~Oza, I.~Hacihaliloglu, and V.~M. Patel, \textit{Medical transformer: Gated axial-attention for medical image segmentation}, \textit{Medical image computing and computer assisted intervention--MICCAI 2021: 24th international conference, Strasbourg, France, September 27--October 1, 2021, proceedings, part I 24}, Springer, 2021, 36--46.

\bibitem{vaswani2017attention}
A.~Vaswani et~al., \textit{Attention is all you need}, Advances in neural information processing systems \textbf{30} (2017).

\bibitem{wang2023lightweight}
L.~Wang, Y.~Yang, A.~Yang, and T.~Li, \textit{Lightweight deep learning model incorporating an attention mechanism and feature fusion for automatic classification of gastric lesions in gastroscopic images}, Biomedical Optics Express \textbf{14} (2023), no.~9, 4677--4695.

\bibitem{wang2018non}
X.~Wang, R.~Girshick, A.~Gupta, and K.~He, \textit{Non-local neural networks}, \textit{Proceedings of the IEEE conference on computer vision and pattern recognition}, 2018, 7794--7803.

\bibitem{xie2021segformer}
E.~Xie, W.~Wang, Z.~Yu, A.~Anandkumar, J.~M. Alvarez, and P.~Luo, \textit{Segformer: Simple and efficient design for semantic segmentation with transformers}, Advances in neural information processing systems \textbf{34} (2021), 12077--12090.

\bibitem{zhao2017pyramid}
H.~Zhao, J.~Shi, X.~Qi, X.~Wang, and J.~Jia, \textit{Pyramid scene parsing network}, \textit{Proceedings of the IEEE conference on computer vision and pattern recognition}, 2017, 2881--2890.

\bibitem{zhou2023nnformer}
H.-Y. Zhou, J.~Guo, Y.~Zhang, X.~Han, L.~Yu, L.~Wang, and Y.~Yu, \textit{nnformer: volumetric medical image segmentation via a 3d transformer}, IEEE transactions on image processing \textbf{32} (2023), 4036--4045.

\bibitem{zhou2018unet++}
Z.~Zhou, M.~M. Rahman~Siddiquee, N.~Tajbakhsh, and J.~Liang, \textit{Unet++: A nested u-net architecture for medical image segmentation}, \textit{Deep learning in medical image analysis and multimodal learning for clinical decision support: 4th international workshop, DLMIA 2018, and 8th international workshop, ML-CDS 2018, held in conjunction with MICCAI 2018, Granada, Spain, September 20, 2018, proceedings 4}, Springer, 2018, 3--11.

\bibitem{zhu2023biformer}
L.~Zhu, X.~Wang, Z.~Ke, W.~Zhang, and R.~W. Lau, \textit{Biformer: Vision transformer with bi-level routing attention}, \textit{Proceedings of the IEEE/CVF conference on computer vision and pattern recognition}, 2023, 10323--10333.

\end{thebibliography}

\end{document}